%% file: main.tex
\documentclass[journal=jctcce,manuscript=article, layout=twocolumn]{achemso}

\usepackage[version=3]{mhchem} 
\usepackage{pgfplots}
\usepackage{tikz}
\usepgfplotslibrary{groupplots}
\pgfplotsset{compat=1.17}
\usepackage{footmisc}
\usepackage{appendix}
\usepackage{graphicx}
\author{Guillem Simeon \footnote{\normalsize{G.S. and A.M. contributed equally to this work.} 
 \label{equalcontrib}}}
\author{Antonio Mirarchi \footref{equalcontrib}}
\author{Raul P. Pelaez}
\affiliation[upf]
{Computational Science Laboratory, Universitat Pompeu Fabra, Barcelona Biomedical Research Park (PRBB), C Dr. Aiguader 88, 08003 Barcelona, Spain.}
\author{Raimondas Galvelis}
\affiliation[upf]
{Computational Science Laboratory, Universitat Pompeu Fabra, Barcelona Biomedical Research Park (PRBB), C Dr. Aiguader 88, 08003 Barcelona, Spain.}
\alsoaffiliation[acellera]{Acellera Labs, C Dr Trueta 183, 08005, Barcelona, Spain}
\author{Gianni De Fabritiis}
\affiliation[upf]
{Computational Science Laboratory, Universitat Pompeu Fabra, Barcelona Biomedical Research Park (PRBB), C Dr. Aiguader 88, 08003 Barcelona, Spain.}
\alsoaffiliation[acellera]{Acellera Labs, C Dr Trueta 183, 08005, Barcelona, Spain}
\alsoaffiliation[icrea]{Instituci\'o Catalana de Recerca i Estudis Avan\c{c}ats (ICREA), Passeig Lluis Companys 23, 08010 Barcelona, Spain}
\email{g.defabritiis@gmail.com}

\usepackage[utf8]{inputenc}
\usepackage[T1]{fontenc}
\usepackage{float}\tolerance=1000
\usepackage{bm}
\usepackage{svg}
\usepackage{amsmath}
\usepackage{graphicx}
\usepackage{float}
\usepackage{amsmath}
\usepackage{amssymb}
\usepackage{hyperref}
\usepackage{color}
\usepackage{enumerate}
\usepackage{listings}
\usepackage{multirow}

\SectionNumbersOn
\renewcommand{\vec}[1]{\bm{#1}}

\date{\today}
\hypersetup{
 pdftitle={TensorNet+},
 pdfkeywords={},
 pdfsubject={},
 pdflang={English}}

\title{Broadening the Scope of Neural Network Potentials through Direct Inclusion of Additional Molecular Attributes}

\begin{document}
\maketitle

\begin{abstract}
Most state-of-the-art neural network potentials do not account for molecular attributes other than atomic numbers and positions, which limits its range of applicability by design. In this work, we demonstrate the importance of including additional electronic attributes in neural network potential representations with a minimal architectural change to TensorNet, a state-of-the-art equivariant model based on Cartesian rank-2 tensor representations. By performing experiments on both custom-made and public benchmarking datasets, we show that this modification resolves the input degeneracy issues stemming from the use of atomic numbers and positions alone, while enhancing the model's predictive accuracy across diverse chemical systems with different charge or spin states. This is accomplished without tailored strategies or the inclusion of physics-based energy terms, while maintaining efficiency and accuracy. These findings should furthermore encourage researchers to train and use models incorporating these additional representations.
\end{abstract}

\section{Introduction}
\label{sec:intro}
Neural network potentials\cite{behlerparrinello, nnp1, ani1, schnet, physnet, nnp4, painn, et, nequip, allegro, mace}, which are deep learning models used to predict the energy and forces of atomic systems\cite{hitchhiker}, have emerged as a tool making possible to circumvent the trade-off between computational cost and accuracy in computational chemistry. Despite its remarkable successes, these architectures typically rely on atomic numbers and positions as their only inputs, resulting in a simplification that overlooks significant quantum properties such as total charge and spin state by design. Neglecting these properties can lead to a degeneracy of inputs, where different molecular states, distinguishable only by their charge or spin, are represented identically by the neural network.

Given these considerations, the quest for extending the range of applicability of these models and for building general neural network potentials requires strategies making the incorporation of these additional attributes possible. For example, as a consequence of these current limitations, researchers have sometimes curated real-world datasets \cite{spice} to avoid the inclusion of charged species \cite{mace-off}, thereby restricting the diversity and realism of the training data, or directly avoided a special treatment of these attributes\cite{schani}. To overcome these challenges, it is essential to integrate these properties \cite{spookynet, aimnet2, bamboo}.

The incorporation of charges has been previously addressed in the literature to take into account the long-range effects that they induce in molecular systems\cite{physnet, cent1, cent2, 4g, Shaidu2024, aimnet-nse, Boittier2024, bamboo}. Previous approaches have typically involved mechanisms for the global redistribution of charge across the system, with neural networks informed by the chemical environment that predict intermediate properties such as electronegativities and effective charges, and equilibration schemes\cite{chargeeq} that solve systems of linear equations\cite{cent1, cent2, 4g, Shaidu2024} or use self-consistent processes\cite{scfnn}. In other models\cite{physnet,spookynet} additional quantities interpreted as atomic charges are predicted and used in ad-hoc introduced physics-based terms for energy prediction, such as Coulomb potential energies. Recently, the Latent Ewald Summation method \cite{les, cacelr} proposed learning hidden charges from local features to compute long-range interactions, avoiding explicit charge equilibration while maintaining accuracy. While this approach can naturally handle long-range effects, an electrostatic-like interaction in reciprocal space is assumed.

In this work, we introduce a simple yet effective extension to a state-of-the-art model such as TensorNet\cite{tensornet}, allowing it to accommodate charged molecules and spin states without requiring architectural changes or additional featurizations, and therefore breaking the aforementioned degeneracy problem that most of the current models display. Despite not explicitly accounting for the global redistribution of properties, this enhancement retains the model's original predictive accuracy without using additional physical terms or predicted quantities, being able to achieve state-of-the-art accuracy in common benchmark datasets. On top of this, since the presented method allows one to include these additional attributes in the learned representations, the prediction of other quantities used in the literature for the incorporation of long-range interactions is facilitated. Overall, the modification of TensorNet presented here expands its applicability to a broader range of chemical systems, addressing a critical shortcoming of most state-of-the-art architectures.

\section{Method}
\label{sec:method}
Given some input atomic numbers and positions $\{Z,\mathbf{r}\}$, TensorNet learns for every atom $(i)$ a set of rank-2 tensors (3x3 matrices) $X^{(i)}$. TensorNet's operations make these representations equivariant to rotations and reflections of the input, meaning that given some matrix $R \in O(3)$, when the matrix is applied on input positions $\{\mathbf{r}\} \rightarrow R\{\mathbf{r}\}$, atomic representations transform as $X^{(i)} \rightarrow RX^{(i)}R^{\mathrm{T}}$, where $R^{\mathrm{T}}$ denotes the transpose of $R$. Furthermore, $X^{(i)}$ can be decomposed into scalar, vector, and symmetric traceless components, $I^{(i)}, A^{(i)}$ and $S^{(i)}$, respectively. After several message-passing layers, the squared Frobenius norms of the representations, which are invariant under $O(3)$, are further processed by the neural network to predict energies, obtaining atomic forces via automatic differentiation.

One of the key operations in TensorNet, distinguishing it from other higher-rank spherical equivariant models, is how neighboring nodes' tensor features are aggregated and used together with the ones of the receiving node to generate a new set of node tensor representations that transform appropriately under $O(3)$. We refer the reader to Ref~\citenum{tensornet} for full details and derivations. In each layer, after some node level transformations $X^{(i)} \rightarrow X'^{(i)} = X^{(i)}/(||X^{(i)}|| + 1) \rightarrow Y^{(i)}$, pair-wise messages $M^{(ij)}$ from neighbor $(j)$ to receiving atom $(i)$ are built by decomposing neighbor's features $Y^{(j)}$,
\begin{equation}
M^{(ij)} = \phi(r_{ij})\big({f_{I}}^{ij} I^{(j)} + {f_{A}}^{ij} A^{(j)} + {f_{S}}^{ij} S^{(j)}\big)
\end{equation}
where ${f_{I}}^{ij}, {f_{A}}^{ij}, {f_{S}}^{ij}$ are learnable functions of the distance between atoms $r_{ij}$, and $\phi(r_{ij})$ is a cosine cutoff function. Tensor messages are summed for all neighbors,
\begin{equation}
    M^{(i)} = \sum_{j\in \mathcal{N}(i)}{M^{(ij)}}
\end{equation}
and the generation of new features $Y'^{(i)}$ from current node features $Y^{(i)}$ and corresponding message $M^{(i)}$ is performed via simple matrix product as
\begin{equation}
Y'^{(i)} = Y^{(i)}M^{(i)} + M^{(i)}Y^{(i)},
\end{equation}
effectively mixing scalars, vectors, and tensors, and ensuring $O(3)$-equivariance, as proved in Ref~\citenum{tensornet}. Resulting representations are eventually manipulated to yield residual updates $\Delta X'^{(i)}$ to the layer's input normalized features $X'^{(i)}$
\begin{equation}
X^{(i)} \leftarrow X'^{(i)} + \Delta X'^{(i)} + (\Delta X'^{(i)})^{2},
\end{equation}
using $X^{(i)}$ to feed the following layer and restart the process.

We propose to include molecular states' information $\psi_k$, by modifying Equations (2) and (3) in TensorNet's interaction layers in the following node-wise manner
\begin{equation}
Y'^{(i)} = (1 + {\lambda_k \psi_k})(Y^{(i)}M^{(i)} + M^{(i)}Y^{(i)}),
\end{equation}
\begin{equation}
X^{(i)} \leftarrow X'^{(i)} + \Delta X'^{(i)} + (1 + {\tilde{\lambda}_{k} \psi_k})(\Delta X'^{(i)})^{2}
\label{eq:lambda}
\end{equation}
where ${\lambda}_{k}, \tilde{\lambda}_{k}$ can be regarded as per-layer constant or learnable weights for the encoding of states such as total charge or spin $\psi_k = \{Q, S, ...\}$. Notice that the modification, while forcing the network to output different values when considering the same input atomic numbers and positions, does not break equivariance under $O(3)$, since it amounts to a rescaling of tensor features by means of a state-dependent scalar factor. Furthermore, it has been designed in such a way that when the system is neutral and in its singlet state, i.e.
\begin{equation}
Q = S = 0,
\end{equation}
the model defaults to the original TensorNet (Eqs (2) and (3)), therefore recovering the predictive accuracy already demonstrated in Ref~\citenum{tensornet}.

Note that since operations are performed node-wise, this framework naturally accommodates both atomic and molecular attributes. For atomic attributes (such as partial charges $\psi_k = {q^{(i)}}$), each node's operations can be modified by its specific value. For molecular attributes (like total charge $\psi_k = Q$ or spin $\psi_k = S$), the same value modifies operations across all nodes in the system. This allows incorporating either local or global information within the same mathematical framework. In the case of atomic charges, commonly used in biomolecular simulations, one needs to rely on some external partial charge computation scheme, which might assume the availability of information beyond atomic numbers and positions, such as SMILES representations, molecular bonds or other preprocessing steps. Furthermore, given the possibility of including total charge as an input to the network, our method also enables the prediction of additional atomic quantities that can be trained to match DFT partial charges and used to evaluate electrostatic terms, as previously done in the literature\cite{spookynet, 4g, Shaidu2024}. Nevertheless, we won't make use of this capability in our experiments, showing that highly accurate models such as TensorNet do not require the inclusion of physics-based terms, such as Coulomb, dispersion or nuclear repulsion, for the benchmark systems studied in this work and commonly used in the literature.

Note that while this framework theoretically supports multiple attributes, careful consideration would be needed to handle them simultaneously. For example, one could consider the modification of Equations 5 and 6, and promote $\lambda_k \psi_k$ to a linear combination of attributes $\sum_{k} \lambda_k \psi_k$. However, fixed weights could lead to degeneracies, where distinct quantum states produce identical scalar factors. A more sophisticated approach using learnable weighting parameters $\lambda_k$ could help discover encodings that properly distinguish different quantum states while preserving their physical influence. The current implementation demonstrates the effectiveness of individual attributes separately, leaving the exploration of simultaneous quantum attributes as future work.

\section{Results and discussion}
\label{sec:results}

We performed a series of experiments with the extended TensorNet model using the TorchMD-Net framework\cite{tmdnet}. We trained the model on a custom-made dataset and several publicly available datasets, targeting atomic number and geometry degeneracy, as well as the simultaneous presence of different charge states regardless of structural or geometric overlap between these. We used a generic set of reasonable hyperparameters, without addressing dataset-dependent fine-tuning to obtain the best possible performances. In all cases, we used the direct model prediction of energies and forces (when needed), without the inclusion of physics-based terms. Overall, we evaluate TensorNet's extension when dealing with total charges, Gasteiger partial charges, or singlet and triplet states. Training details and hyperparameters used in the experiments can be found in the Supplementary Information.

\subsection{Toy degeneracy problem}

\begin{figure*}
    \centering
    \includegraphics[width=0.8\textwidth]{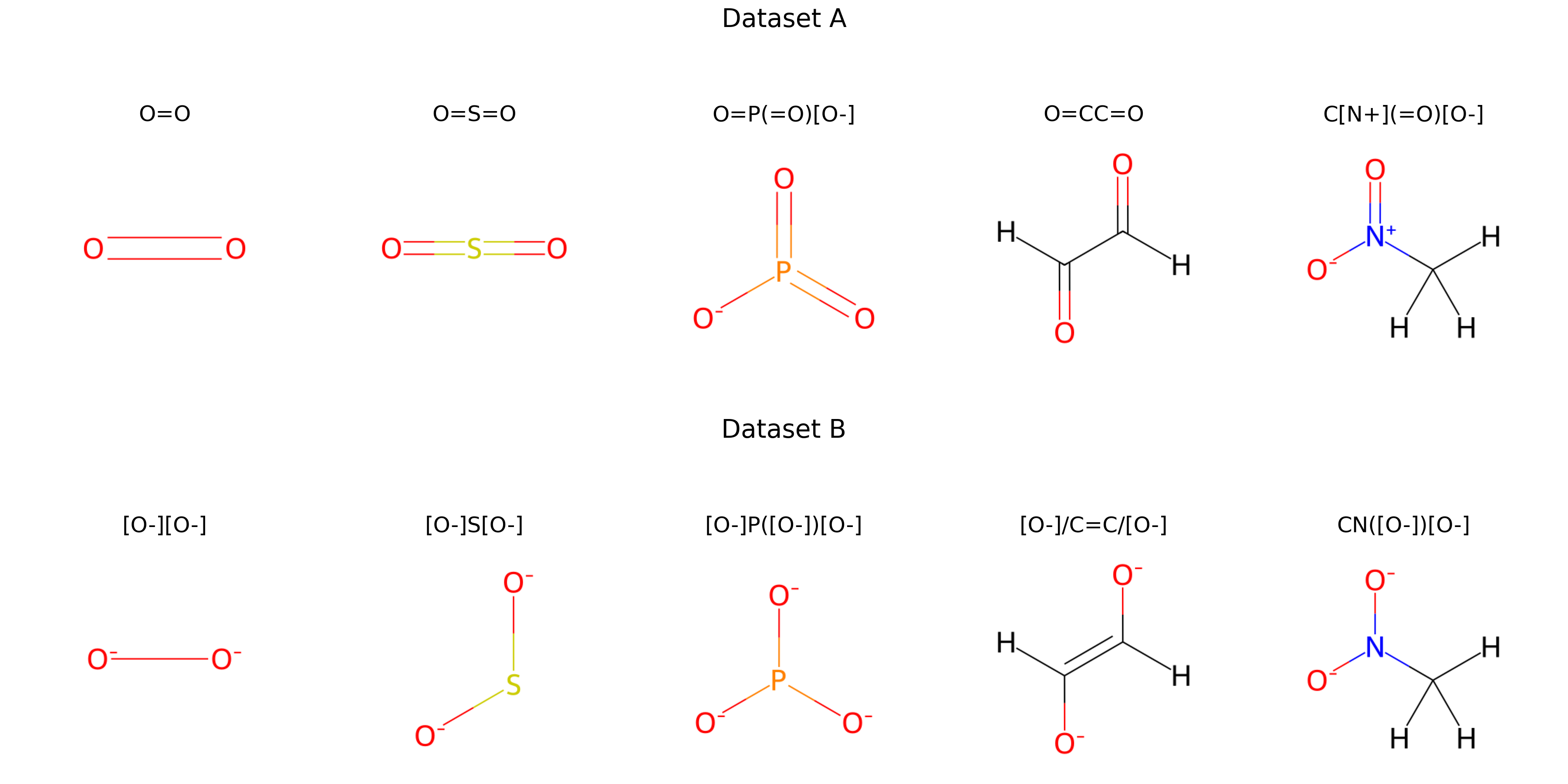}
    \caption{Molecules included in the A and B toy datasets, from which 2,000 data points per molecule are obtained by generating conformers and computing potential energies and atomic forces using GFN2-xTB\cite{xtb}. Columns illustrate degenerate pairs of molecules for a neural network that uses solely atomic numbers and positions as inputs.}
    \label{fig:toy-dataset-fig}
\end{figure*}

As a first test, to illustrate the input degeneracy issue and how TensorNet's extension resolves it without affecting its baseline accuracy, we constructed two toy datasets, Dataset A and Dataset B, each one comprising five members of five pairs of unique molecular systems, each of the elements in the pair differing in total charge (see Fig \ref{fig:toy-dataset-fig}). These pairs are indistinguishable from a neural network that does not account for total charge, as they share identical atomic numbers and geometric configurations. The datasets include total charges, Gasteiger partial charges computed with RDKit, and calculated energies and atomic forces for 2000 conformers per molecule using GFN2-xTB\cite{xtb}. Therefore, each dataset contains a total of 10k data points. Conformers were generated by minimizing each molecule, displacing atomic positions with Gaussian noise with a standard deviation of 0.2 \AA, and filtering them such that maximum atomic forces are <100 eV/\AA.

The logic behind the experiment is the following: when training the original TensorNet on each dataset separately, the learning should proceed as expected, with the network successfully learning the mapping between the atomic inputs and the output properties. However, when the datasets are combined, the inability to distinguish charge states leads to an overlap in the input space, being mapped to different values of energies and forces. As a result, the network should fail to learn accurately. On the other hand, the extension of the model should allow the network to learn accurately the mapping between inputs and outputs in all cases.

Splits of 50/10/40\% were used for training, validation, and testing, respectively, both for single and combined datasets. For the charge-aware case, we tried separately the inclusion of total charges $Q$ and Gasteiger partial charges $q^{(i)}$, both with $\lambda_Q = \tilde{\lambda}_{Q}$ and $\lambda_{q^{(i)}} = \tilde{\lambda}_{{q^{(i)}}}$ equal to 0.1 across all layers. Results can be found in Table \ref{tab:toy}. As expected, the original model trained on the merged dataset has a very poor accuracy (TensorNet, Dataset A $\cup$ B). The extension with total and partial charges allows us to learn the merged dataset with sub-meV and sub-meV/\r{A} differences in energy and force errors with respect to single dataset training of the corresponding extended model. The use of total charge $Q$ gives better results than the use of partial charges $q^{(i)}$.
\begin{table*}[ht!]
\centering
\caption{\textbf{Toy datasets results.} Energy (E) and forces (F) mean absolute errors in meV and meV/\r{A}. $Q$ refers to total charge while $q^{(i)}$ refers to Gasteiger partial charges, with factors corresponding to weights $\lambda_Q = \tilde{\lambda}_{Q}$ and $\lambda_{q^{(i)}} = \tilde{\lambda}_{{q^{(i)}}}$, respectively. 
}
\begin{tabular}{ccccc}
\hline \\[-2ex]
\textbf{Model} &  & \textbf{Dataset A} & \textbf{Dataset B} & \textbf{Dataset A $\cup$ B}\\ \hline
 \multirow{2}{*}{TensorNet [\citenum{tensornet}]} & E (meV) & 2.6 & 2.3 & 4437 \\
& F (meV/\r{A})& 11.6 & 14.3 & 772 \\ \hline
 \multirow{2}{*}{TensorNet+$0.1Q$} & E (meV)& 2.4 & 2.3 & 2.5 \\
& F (meV/\r{A})& 11.9 & 14.8 & 13.9 \\ \hline
\multirow{2}{*}{TensorNet+$0.1q^{(i)}$} & E (meV)& 2.8 & 2.3 & 3.2\\
& F (meV/\r{A})& 13.7 & 14.9 & 14.8\\ \hline
\end{tabular}
\label{tab:toy}
\end{table*}

\subsection{SPICE PubChem}

The PubChem subset inside the SPICE dataset\cite{spice} (version 1.1.4) consists of conformers of small drug-like molecules, with energies and forces computed at the $\omega$B97M-D3(BJ)/def2-TZVPPD level of theory. Approximately 4\% of the molecules and conformers in the dataset are not neutral. The dataset already provides total charges, while we computed Gasteiger partial charges for all the molecules in the subset, including the neutral ones. The computation of partial charges failed with RDKit for 42 molecules, with corresponding conformers being discarded. This resulted in 707,558 data points in total. To include total charge $Q$ or atomic partial charges $q^{(i)}$, we take $\lambda_Q = \tilde{\lambda}_{Q}$ and $\lambda_{q^{(i)}} = \tilde{\lambda}_{{q^{(i)}}}$ constant and non-learnable across all layers, equal to 0.1 or 0.25. We used 80/10/10\% splits for training, validation, and testing, respectively. 

In Table 2, we show that including total charge $Q$ or partial charges $q^{(i)}$ can improve energy and forces accuracy. Even though charged species comprise only 4\% of the data points, the freedom given to the model to adapt to varying charge states improves significantly the accuracy on both charged and neutral conformations. With the best method, which uses total charge $Q$ with $\lambda_Q = \tilde{\lambda}_{Q} = 0.1$, reductions of 30\% and 12\% in energy errors are achieved for charged and neutral test molecules, respectively. The persisting performance gap between neutral and charged molecules (Table 2) can be attributed to dataset imbalance. This is evidenced by our toy dataset experiments (Table 1), where an equal representation of charge states (2000 conformations each) shows no accuracy degradation for the merged dataset. In fact, force prediction accuracy for the merged dataset falls between that of individual datasets, indicating comparable performance across charge states when properly balanced. The imbalanced case can be understood as model parameters being predominantly trained on $Q=0$ data, with a limited representation of other charge values. Nevertheless, even with just 4\% charged species, including total charge consistently improves overall and charge-specific accuracy. This consistent improvement in prediction accuracy for both neutral and charged species suggests that the development of distinct internal representations due to charge awareness benefits the model beyond just handling charged molecules. This demonstrates that incorporating attributes like total charge helps the model develop more physically meaningful representations, enhancing predictive performance across all molecular states. The inclusion of Gasteiger partial charges gives relevant out-of-the-box improvements too.

\begin{table*}[ht]
\centering
\caption{\textbf{SPICE PubChem results.} Energy (E) and forces (F) mean absolute errors in meV and meV/\r{A}, on the entire test set, and separately on neutral and charged conformations in the test set. $Q$ refers to total charge while $q^{(i)}$ refers to Gasteiger partial charges, with factors corresponding to weights $\lambda_Q = \tilde{\lambda}_{Q}$ and $\lambda_{q^{(i)}} = \tilde{\lambda}_{{q^{(i)}}}$, respectively. 
}
\begin{tabular}{ccccc}
\hline \\[-2ex]
\textbf{Model} &  & \textbf{Test total} & \textbf{Test neutral} & \textbf{Test charged}\\ \hline
 \multirow{2}{*}{TensorNet [\citenum{tensornet}]} & E (meV) & 34.2 & 32.7 & 70.1 \\
& F (meV/\r{A})& 34.7 & 33.8 & 58.4\\ \hline
 \multirow{2}{*}{TensorNet+$0.1q^{(i)}$} & E (meV) & 31.5 & 30.5 & 55.2 \\
& F (meV/\r{A}) & 33.9 & 33.0 & 56.7\\ \hline
\multirow{2}{*}{TensorNet+$0.25q^{(i)}$} & E (meV) & 30.5 & 29.6 & 50.4\\
& F (meV/\r{A}) & 32.9 & 32.1 & 55.6 \\ \hline
\multirow{2}{*}{TensorNet+$0.1Q$} & E (meV) & \textbf{29.4} & \textbf{28.6} & \textbf{49.3}\\
& F (meV/\r{A}) & \textbf{31.4} & \textbf{30.6} & \textbf{50.1}\\ \hline
\multirow{2}{*}{TensorNet+$0.25Q$} & E (meV) & 29.6 & \textbf{28.6} & 52.0\\
& F (meV/\r{A}) & 31.6 & \textbf{30.6} & 52.9\\ \hline
\end{tabular}
\label{tab:spice}
\end{table*}

\subsection{Carbon chain, silver clusters and sodium chloride clusters}
These datasets were introduced in Ref~\citenum{4g} to illustrate the shortcomings of previous generations of models \cite{3g1, 3g2}, and are typically used as benchmarks for architectures incorporating long-range interactions and electrostatics, allowing us to compare their performances against the modified TensorNet, which does not incorporate physics-based terms in the energy prediction.
\newline
\textbf{Carbon chain}, $\mathbf{C_{10}H_{2}/C_{10}H_{3}^{+}}$. This benchmark consists of two subsystems: a neutral linear carbon chain with hydrogen atoms at each end, $\mathrm{C_{10}H_{2}}$, and a protonated counterpart, $\mathrm{C_{10}H_{3}^{+}}$, yielding a total charge of +1 and resulting in a global charge redistribution.
\newline
\textbf{Silver clusters}, $\mathbf{Ag_{3}^{+/-}}$. In this case, the dataset consists of triangular and linear Ag trimers with total charges of +1 and –1, respectively. These systems define an ill-posed learning problem for models which are unaware of charge in a similar way to the toy datasets from Section 3.1. 
\newline
\textbf{Sodium chloride clusters}, $\mathbf{Na_{8/9}Cl_{8}^{+}}$. This benchmark contains a positively charged $\mathrm{Na_{9}Cl_{8}^{+}}$ cluster and the system resulting from the removal of a neutral sodium atom, $\mathrm{Na_{8}Cl_{8}^{+}}$. The +1 total charge of the cluster is maintained, resulting in a redistribution of charge when the sodium atom is removed.

In line with previous work, we trained extended TensorNet on 90\% data points, being the remaining ones used as a separate validation set. Comparative results for 4D-HDNNP\cite{4g}, PANNA-LR\cite{Shaidu2024}, CACE-LR \cite{cacelr}, SpookyNet\cite{spookynet} and TensorNet can be found in Table \ref{tab:bench}. In the case of 4G-HDNNP, PANNA-LR ($\gamma_q > 0$) and SpookyNet, the models are trained to reproduce DFT atomic reference charges and include an explicit electrostatic term in the energy prediction that makes use of these. In SpookyNet, nuclear repulsion and D4 dispersion terms are also included. We emphasize that, in contrast, TensorNet was directly trained on energy and force labels without making use of any charge predictions or prior physical terms. CACE-LR makes use of Latent Ewald Summation \cite{les}, which learns hidden 'latent charges' from atomic features without explicit reference to any specific charge definition, and uses these to compute long-range interactions via Ewald summation while avoiding the need for charge equilibration. However, one can consider the assumed interaction in reciprocal space as an electrostatic-like ansatz. Therefore, we compared our approach to a wide diversity of methods in the literature.

Our model performs competitively or outperforms SpookyNet, the current best performing architecture. The baseline TensorNet already shows strong performance on the $\mathrm{C_{10}H_{2}/C_{10}H_{3}^{+}}$ and $\mathrm{Na_{8/9}Cl_{8}^{+}}$ systems even without charge awareness. This is because these systems can be distinguished structurally: $\mathrm{C_{10}H_{3}^{+}}$ has an additional proton compared to $\mathrm{C_{10}H_{2}}$, while $\mathrm{Na_{9}Cl_{8}^{+}}$ has one more sodium atom than $\mathrm{Na_{8}Cl_{8}^{+}}$. However, for $\mathrm{Ag_{3}^{+/-}}$, where three silver atoms form either a cation or anion, the baseline model fails to distinguish between charge states, leading to large errors (1683.11 meV/atom). The inclusion of charge awareness maintains the strong performance on structurally distinguishable systems while dramatically improving predictions for $\mathrm{Ag_{3}^{+/-}}$ (0.888 meV/atom), demonstrating its ability to resolve input degeneracy when geometric information alone is insufficient.

Furthermore, we compared the speed performance of SpookyNet and TensorNet for a carbon chain conformation, using the ASE Calculator implementation \cite{ase-paper, ase-ref-2} for SpookyNet and the TorchMD-Net external calculator for TensorNet. The speed benchmark was conducted on an NVIDIA RTX 4090, with both models calculating energy and forces over 100 runs. TensorNet achieved a mean runtime of 1.462 ms per inference step, while SpookyNet recorded 1.751 ms. Notably, while SpookyNet's long-range electrostatics scale as $O(N^2)$, our method scales as $O(N)$.

\begin{table*}[ht]
\centering
\caption{\textbf{Charged benchmark systems} $\mathbf{C_{10}H_{2}/C_{10}H_{3}^{+}}$, $\mathbf{Ag_{3}^{+/-}}$ \textbf{and} $\mathbf{Na_{8/9}Cl_{8}^{+}}$. Energy (E) and forces (F) root mean squared errors in meV/atom and meV/\r{A}, with standard deviation over 3 splits between parentheses for TensorNet, on the validation set consisting of 10\% of data points. $Q$ refers to total charge, with factor corresponding to weights $\lambda_Q = \tilde{\lambda}_{Q} = 0.1$. Results for 4G-HDNNP and SpookyNet have been taken from ~\citenum{spookynet}, PANNA-LR results have been extracted from ~\citenum{Shaidu2024} corresponding to the best-performing model trained on partial charges, $\gamma_q > 0$ as denoted in the reference, while CACE-LR results have been obtained from ~\citenum{cacelr}. 
}
\begin{tabular}{ccccc}
\hline \\[-2ex]
\textbf{Model} &  & $\mathbf{C_{10}H_{2}/C_{10}H_{3}^{+}}$ & $\mathbf{Ag_{3}^{+/-}}$ & $\mathbf{Na_{8/9}Cl_{8}^{+}}$\\ \hline
 \multirow{2}{*}{4G-HDNNP [\citenum{4g}]} & E (meV/atom) & 1.194 & 1.323 & 0.481 \\
& F (meV/\r{A})& 78.00 & 31.69 & 32.78 \\ \hline
\multirow{2}{*}{PANNA-LR [\citenum{Shaidu2024}]} & E (meV/atom) & 1.17 & 0.80 & 0.40 \\
& F (meV/\r{A}) & 79 & \textbf{20} & 19 \\ \hline
\multirow{2}{*}{CACE-LR [\citenum{cacelr}]} & E (meV/atom) & 0.73 & \textbf{0.162} & 0.21 \\
& F (meV/\r{A}) & 36.9 & 29.0 & 9.78 \\ \hline
\multirow{2}{*}{SpookyNet [\citenum{spookynet}]} & E (meV/atom) & 0.364 & 0.220 & 0.135 \\
& F (meV/\r{A}) & 5.802 & 26.64 & \textbf{1.052} \\ \hline
\multirow{2}{*}{TensorNet} & E (meV/atom) & \textbf{0.124} (0.001) & 1683.11 (203.36)  & \textbf{0.103} (0.003)\\
& F (meV/\r{A}) & \textbf{1.88} (0.02) & 912.67 (39.20) & 1.73 (0.06) \\
\hline
\multirow{2}{*}{TensorNet+$0.1Q$} & E (meV/atom) & \textbf{0.124} (0.001) & 0.888 (0.019) & \textbf{0.102} (0.003)\\
& F (meV/\r{A}) & \textbf{1.90} (0.15) & \textbf{20.70} (0.77) & 1.78 (0.09)\\ 
\hline
\end{tabular}
\label{tab:bench}
\end{table*}

\subsection{Solvated protein fragments}

The solvated protein fragments dataset, introduced in Ref~\citenum{physnet}, includes structures for all possible hydrogen-saturated covalently bonded fragments with up to eight heavy atoms (C, N, O, S) derived from proteins. It accounts for different charge states of aminoacids due to  (de)protonation, with total charges up to ±2. The dataset features solvated variants with varying water molecule numbers, and randomly sampled dimer interactions of protein fragments, along with pure water structures with up to 40 molecules. Multiple conformations are included, resulting in more than 2.7M structures with reference energies, forces, and dipole moments.

In line with Ref ~\citenum{physnet}, we trained TensorNet with the inclusion of total charge ($\lambda_Q = \tilde{\lambda}_{Q} = 0.1$) on the energy and forces, excluding dipoles, of 2560k structures, with 100k of these being used for validation, being the remaining ones used for testing. Results are found in Table \ref{tab:solv}, and demonstrate that the modified TensorNet significantly outperforms the more accurate ensemble of five PhysNet models, which use explicit electrostatic and dispersion energy terms in the predictions, particularly in energies, nearly halving its error.

\begin{table*}[ht]
\centering
\caption{\textbf{Solvated protein fragments results.} Comparison of mean absolute errors on the solvated protein fragments test set for energies (E) and forces (F). $Q$ refers to total charge, with the factor corresponding to $\lambda_Q = \tilde{\lambda}_{Q} = 0.1$. PhysNet-ens5 denotes an ensemble of five PhysNet models. Results have been taken from ~\citenum{physnet}.}
\begin{tabular}{ccccc}
\hline \\[-2ex]
& \textbf{PhysNet} & \textbf{PhysNet-ens5} & \textbf{TensorNet+$0.1Q$} \\
[0.3ex] \hline
\\[-2ex]
E & 1.03 kcal/mol & 0.95 kcal/mol & \textbf{0.55} (0.04) kcal/mol\\
[0.3ex] \hline
\\[-2ex]
F & 0.88 kcal/mol/\r{A} & 0.72 kcal/mol/\r{A} & \textbf{0.63} (0.09) kcal/mol/\r{A}\\[0.2ex]\hline
\end{tabular}
\label{tab:solv}
\end{table*}

\subsection{QMspin}

The QMspin dataset, introduced in Ref~\citenum{qmspin}, includes molecules optimized in both singlet and triplet states. The dataset, drawn from the QM9 database, provides a comprehensive collection of around 13,000 carbene structures, with energies calculated for both singlet and triplet states. Therefore, QMspin contains approximately 26,000 data points. The singlet and triplet energy information enables the assessment of the model's predictive performance on spin state differences, an aspect previously unaddressed due to the degeneracy of inputs, allowing us to test and validate TensorNet's extension beyond charged states. We incorporate singlet or triplet states as $S = 0$ and $S = 1$, respectively, and $\lambda_S = \tilde{\lambda}_{S} = 0.1$ non-learnable for all layers.

We removed 228 data points, which corresponded to geometry files where the number of atoms in the header did not match the number of atoms with coordinate entries. Following Ref~\citenum{spookynet}, we used 20k and 1k data points for training and validation, respectively, the remaining data points being used for testing. The results, found in Table \ref{tab:qmspin}, show that the inclusion of the spin state $S$, and therefore degeneracy breaking, improves 10-fold the accuracy of the model, achieving 43 meV error in energies, considered chemical accuracy. Furthermore, this represents an improvement of $\sim$40\% with respect to SpookyNet\cite{spookynet}, the only model that to the best of our knowledge has been benchmarked against the QMspin dataset. Again, SpookyNet models explicitly electrostatic, dispersion, and nuclear repulsion interactions.

\begin{table*}[ht]
\centering
\caption{\textbf{QMspin results.} Comparison of mean absolute errors on the QMspin test set energies (E), in meV. $S$ refers to spin state, with the factor corresponding to $\lambda_S = \tilde{\lambda}_{S} = 0.1$.}
\begin{tabular}{cccc}
\hline \\[-2ex]
 & \textbf{SpookyNet} [\citenum{spookynet}] & \textbf{TensorNet} &\textbf{TensorNet+$0.1S$}\\
[0.3ex] \hline
\\[-2ex]
E & 68 meV & 432 (19) meV &\textbf{43} (2) meV\\[0.2ex]\hline
\end{tabular}
\label{tab:qmspin}
\end{table*}

\section{Conclusion}
In this work, we have demonstrated and addressed a significant gap in some state-of-the-art neural network potentials, which typically disregard charge and spin states in the model's representations. We modify TensorNet to accommodate these features with a zero-cost architectural change which effectively resolves degeneracy issues arising from the omission of these critical quantum attributes. Moreover, through experiments on well-established and custom-made benchmark datasets, we have shown that the predictive accuracy of the model is improved and that it performs on par with or outperforms existing methods that require more complex treatments of charges and spins.

Our results highlight that even a straightforward incorporation of these properties can lead to substantial improvements without the need for the introduction of additional physical terms or predicted quantities, despite facilitating and not precluding their use. Future work could explore more sophisticated strategies for scaling additional attributes, such as making them learnable scalars dependent on atomic properties.

Although this work focuses mainly on global attributes like total charge and spin, our framework mathematically supports atomic-level attributes like partial charges that could better represent local chemical environments in more complex and heterogeneous systems. However, this capability would require thorough validation and benchmarking beyond the scope of this work. Similarly, though our model effectively handles charge effects through implicit learning, it does not guarantee the correct asymptotic behavior of electrostatic interactions that physical laws demand, trading exact long-range physics for architectural simplicity while maintaining strong performance within typical molecular scales.

Overall, given the strong performance exhibited in our results, this work emphasizes the importance of incorporating charge and spin states to achieve high predictive accuracy and generality in modeling chemical systems.

\section*{Data and software availability}

TensorNet and its extension can be found within TorchMD-Net: \url{https://github.com/torchmd/torchmd-net}. The toy datasets have been made available at \url{https://zenodo.org/records/10852523}. The SPICE dataset version 1.1.4 is publicly available at \url{https://zenodo.org/records/8222043}. The QMspin dataset is available at \url{https://archive.materialscloud.org/record/2020.0051/v1}.

\section*{Acknowledgments}
G. S. is financially supported by Generalitat de Catalunya's Agency for Management of University and Research Grants (AGAUR) PhD grant FI-2-00587. A. M. is financially supported by AGAUR PhD grant 2024 FI-1-00278. Research reported in this publication was supported by the project PID2020-116564GB-I00, has been funded by MICIU/AEI /10.13039/501100011033, and by the National Institute of General Medical Sciences (NIGMS) of the National Institutes of Health under award number R01GM140090. 
The content is solely the responsibility of the authors and does not necessarily represent the official views of the National Institutes of Health.
\section*{ASSOCIATED CONTENT}
Supporting Information with further details on training and hyperparameters used for the different experiments presented throughout the manuscript is available.
\bibliography{references}
\input{supplementary_information}
\newpage

\end{document}

%% file: supplementary_information.tex
\renewcommand{\vec}[1]{\bm{#1}}
\setcounter{section}{0} 
\twocolumn[
    \centering
    {\LARGE Supplementary Information: Broadening the Scope of Neural Network Potentials through Direct Inclusion of Additional Molecular Attributes} 
    
]

\twocolumn

\section{Hyperparameters}
We include hyperparameters used to generate the different trainings presented throughout the manuscript.
Experiments were performed with the TorchMD-Net framework on two NVIDIA RTX 4090, except for Solvated Protein Fragments which was performed on four, using PyTorch Lightning's DDP multi-GPU training protocol. 

\begin{table}
\centering
\caption{\textbf{Toy datasets training details and hyperparameters.}}

\begin{tabular}{lc}
\hline
\textbf{Parameter} & \textbf{Value} \\
\hline
\texttt{activation} & \texttt{silu} \\
\texttt{batch\_size} & \texttt{16} \\
\texttt{cutoff\_lower} & \texttt{0.0} \\
\texttt{cutoff\_upper} & \texttt{5.0} \\
\texttt{derivative} & \texttt{True} \\
\texttt{early\_stopping\_patience} & \texttt{100} \\
\texttt{embedding\_dimension} & \texttt{128} \\
\texttt{equivariance\_invariance\_group} & \texttt{O(3)} \\
\texttt{gradient\_clipping} & \texttt{40} \\
\texttt{lr} & \texttt{1e-3} \\
\texttt{lr\_factor} & \texttt{0.5} \\
\texttt{lr\_min} & \texttt{1e-7} \\
\texttt{lr\_patience} & \texttt{15} \\
\texttt{lr\_warmup\_steps} & \texttt{0} \\
\texttt{neg\_dy\_weight} & \texttt{10.0} \\
\texttt{num\_layers} & \texttt{2} \\
\texttt{num\_rbf} & \texttt{32} \\
\texttt{seed} & \texttt{1} \\
\texttt{train\_size} & \texttt{0.5} \\
\texttt{val\_size} & \texttt{0.1} \\
\texttt{y\_weight} & \texttt{1.0} \\
\hline
\end{tabular}
\end{table}

\begin{table}
\centering
\caption{\textbf{SPICE PubChem training details and hyperparameters.}}

\begin{tabular}{lc}
\hline
\textbf{Parameter} & \textbf{Value} \\
\hline
\texttt{activation} & \texttt{silu} \\
\texttt{batch\_size} & \texttt{64} \\
\texttt{cutoff\_lower} & \texttt{0.0} \\
\texttt{cutoff\_upper} & \texttt{5.0} \\
\texttt{derivative} & \texttt{True} \\
\texttt{early\_stopping\_patience} & \texttt{50} \\
\texttt{embedding\_dimension} & \texttt{128} \\
\texttt{equivariance\_invariance\_group} & \texttt{O(3)} \\
\texttt{gradient\_clipping} & \texttt{100} \\
\texttt{lr} & \texttt{1e-3} \\
\texttt{lr\_factor} & \texttt{0.5} \\
\texttt{lr\_min} & \texttt{1e-7} \\
\texttt{lr\_patience} & \texttt{5} \\
\texttt{lr\_warmup\_steps} & \texttt{0} \\
\texttt{neg\_dy\_weight} & \texttt{10.0} \\
\texttt{num\_layers} & \texttt{2} \\
\texttt{num\_rbf} & \texttt{32} \\
\texttt{seed} & \texttt{1} \\
\texttt{train\_size} & \texttt{0.8} \\
\texttt{val\_size} & \texttt{0.1} \\
\texttt{y\_weight} & \texttt{1.0} \\
\hline
\end{tabular}
\end{table}

\begin{table}
\centering
\caption{\textbf{Carbon chain, silver clusters, and sodium chloride clusters training details and hyperparameters.} In this case, a linear fit was performed beforehand on the training set to obtain element-wise reference energies, which were subtracted from total energy labels. For $\mathrm{Ag_{3}^{+/-}}$, \texttt{y\_weight} was 10.0.}

\begin{tabular}{lc}
\hline
\textbf{Parameter} & \textbf{Value} \\
\hline
\texttt{activation} & \texttt{silu} \\
\texttt{batch\_size} & \texttt{4} \\
\texttt{cutoff\_lower} & \texttt{0.0} \\
\texttt{cutoff\_upper} & \texttt{6.0} \\
\texttt{derivative} & \texttt{True} \\
\texttt{early\_stopping\_patience} & \texttt{300} \\
\texttt{embedding\_dimension} & \texttt{128} \\
\texttt{equivariance\_invariance\_group} & \texttt{O(3)} \\
\texttt{gradient\_clipping} & \texttt{100} \\
\texttt{lr} & \texttt{1e-3} \\
\texttt{lr\_factor} & \texttt{0.5} \\
\texttt{lr\_min} & \texttt{1e-7} \\
\texttt{lr\_patience} & \texttt{15} \\
\texttt{lr\_warmup\_steps} & \texttt{100} \\
\texttt{neg\_dy\_weight} & \texttt{1.0} \\
\texttt{num\_layers} & \texttt{2} \\
\texttt{num\_rbf} & \texttt{32} \\
\texttt{seed} & \texttt{1} \\
\texttt{train\_size} & \texttt{0.9} \\
\texttt{val\_size} & \texttt{0.1} \\
\texttt{y\_weight} & \texttt{1.0} \\
\hline
\end{tabular}
\end{table}

\begin{table}
\centering
\caption{\textbf{Solvated protein fragments training details and hyperparameters.}}

\begin{tabular}{lc}
\hline
\textbf{Parameter} & \textbf{Value} \\
\hline
\texttt{activation} & \texttt{silu} \\
\texttt{batch\_size} & \texttt{32} \\
\texttt{cutoff\_lower} & \texttt{0.0} \\
\texttt{cutoff\_upper} & \texttt{5.0} \\
\texttt{derivative} & \texttt{True} \\
\texttt{early\_stopping\_patience} & \texttt{50} \\
\texttt{embedding\_dimension} & \texttt{128} \\
\texttt{equivariance\_invariance\_group} & \texttt{O(3)} \\
\texttt{gradient\_clipping} & \texttt{100} \\
\texttt{lr} & \texttt{1e-3} \\
\texttt{lr\_factor} & \texttt{0.5} \\
\texttt{lr\_min} & \texttt{1e-7} \\
\texttt{lr\_patience} & \texttt{10} \\
\texttt{lr\_warmup\_steps} & \texttt{0} \\
\texttt{neg\_dy\_weight} & \texttt{10.0} \\
\texttt{num\_layers} & \texttt{2} \\
\texttt{num\_rbf} & \texttt{32} \\
\texttt{seed} & \texttt{1} \\
\texttt{train\_size} & \texttt{2560000} \\
\texttt{val\_size} & \texttt{100000} \\
\texttt{y\_weight} & \texttt{1.0} \\
\hline
\end{tabular}
\end{table}

\begin{table}
\centering
\caption{\textbf{QMspin training details and hyperparameters.} In this case, a linear fit was performed beforehand on the training set to obtain element-wise reference energies, which were subtracted from total energy labels.}

\begin{tabular}{lc}
\hline
\textbf{Parameter} & \textbf{Value} \\
\hline
\texttt{activation} & \texttt{silu} \\
\texttt{batch\_size} & \texttt{16} \\
\texttt{cutoff\_lower} & \texttt{0.0} \\
\texttt{cutoff\_upper} & \texttt{5.0} \\
\texttt{derivative} & \texttt{False} \\
\texttt{early\_stopping\_patience} & \texttt{100} \\
\texttt{embedding\_dimension} & \texttt{128} \\
\texttt{equivariance\_invariance\_group} & \texttt{O(3)} \\
\texttt{gradient\_clipping} & \texttt{40} \\
\texttt{lr} & \texttt{1e-3} \\
\texttt{lr\_factor} & \texttt{0.5} \\
\texttt{lr\_min} & \texttt{1e-7} \\
\texttt{lr\_patience} & \texttt{30} \\
\texttt{lr\_warmup\_steps} & \texttt{0} \\
\texttt{neg\_dy\_weight} & \texttt{0.0} \\
\texttt{num\_layers} & \texttt{2} \\
\texttt{num\_rbf} & \texttt{32} \\
\texttt{seed} & \texttt{1} \\
\texttt{train\_size} & \texttt{20000} \\
\texttt{val\_size} & \texttt{1000} \\
\texttt{y\_weight} & \texttt{1.0} \\
\hline
\end{tabular}
\end{table}

%% file: main.bbl
\providecommand{\latin}[1]{#1}
\makeatletter
\providecommand{\doi}
  {\begingroup\let\do\@makeother\dospecials
  \catcode`\{=1 \catcode`\}=2 \doi@aux}
\providecommand{\doi@aux}[1]{\endgroup\texttt{#1}}
\makeatother
\providecommand*\mcitethebibliography{\thebibliography}
\csname @ifundefined\endcsname{endmcitethebibliography}  {\let\endmcitethebibliography\endthebibliography}{}
\begin{mcitethebibliography}{37}
\providecommand*\natexlab[1]{#1}
\providecommand*\mciteSetBstSublistMode[1]{}
\providecommand*\mciteSetBstMaxWidthForm[2]{}
\providecommand*\mciteBstWouldAddEndPuncttrue
  {\def\EndOfBibitem{\unskip.}}
\providecommand*\mciteBstWouldAddEndPunctfalse
  {\let\EndOfBibitem\relax}
\providecommand*\mciteSetBstMidEndSepPunct[3]{}
\providecommand*\mciteSetBstSublistLabelBeginEnd[3]{}
\providecommand*\EndOfBibitem{}
\mciteSetBstSublistMode{f}
\mciteSetBstMaxWidthForm{subitem}{(\alph{mcitesubitemcount})}
\mciteSetBstSublistLabelBeginEnd
  {\mcitemaxwidthsubitemform\space}
  {\relax}
  {\relax}

\bibitem[Behler and Parrinello(2007)Behler, and Parrinello]{behlerparrinello}
Behler,~J.; Parrinello,~M. Generalized Neural-Network Representation of High-Dimensional Potential-Energy Surfaces. \emph{Phys. Rev. Lett.} \textbf{2007}, \emph{98}, 146401\relax
\mciteBstWouldAddEndPuncttrue
\mciteSetBstMidEndSepPunct{\mcitedefaultmidpunct}
{\mcitedefaultendpunct}{\mcitedefaultseppunct}\relax
\EndOfBibitem
\bibitem[Kocer \latin{et~al.}(2021)Kocer, Ko, and Behler]{nnp1}
Kocer,~E.; Ko,~T.~W.; Behler,~J. Neural Network Potentials: A Concise Overview of Methods. 2021\relax
\mciteBstWouldAddEndPuncttrue
\mciteSetBstMidEndSepPunct{\mcitedefaultmidpunct}
{\mcitedefaultendpunct}{\mcitedefaultseppunct}\relax
\EndOfBibitem
\bibitem[Smith \latin{et~al.}(2017)Smith, Isayev, and Roitberg]{ani1}
Smith,~J.~S.; Isayev,~O.; Roitberg,~A.~E. {ANI}-1: an extensible neural network potential with {DFT} accuracy at force field computational cost. \emph{Chemical Science} \textbf{2017}, \emph{8}, 3192--3203\relax
\mciteBstWouldAddEndPuncttrue
\mciteSetBstMidEndSepPunct{\mcitedefaultmidpunct}
{\mcitedefaultendpunct}{\mcitedefaultseppunct}\relax
\EndOfBibitem
\bibitem[Sch\"{u}tt \latin{et~al.}(2018)Sch\"{u}tt, Sauceda, Kindermans, Tkatchenko, and M\"{u}ller]{schnet}
Sch\"{u}tt,~K.~T.; Sauceda,~H.~E.; Kindermans,~P.-J.; Tkatchenko,~A.; M\"{u}ller,~K.-R. SchNet – A deep learning architecture for molecules and materials. \emph{The Journal of Chemical Physics} \textbf{2018}, \emph{148}\relax
\mciteBstWouldAddEndPuncttrue
\mciteSetBstMidEndSepPunct{\mcitedefaultmidpunct}
{\mcitedefaultendpunct}{\mcitedefaultseppunct}\relax
\EndOfBibitem
\bibitem[Unke and Meuwly(2019)Unke, and Meuwly]{physnet}
Unke,~O.~T.; Meuwly,~M. PhysNet: A Neural Network for Predicting Energies, Forces, Dipole Moments, and Partial Charges. \emph{Journal of Chemical Theory and Computation} \textbf{2019}, \emph{15}, 3678–3693\relax
\mciteBstWouldAddEndPuncttrue
\mciteSetBstMidEndSepPunct{\mcitedefaultmidpunct}
{\mcitedefaultendpunct}{\mcitedefaultseppunct}\relax
\EndOfBibitem
\bibitem[Deringer \latin{et~al.}(2019)Deringer, Caro, and Csányi]{nnp4}
Deringer,~V.~L.; Caro,~M.~A.; Csányi,~G. Machine Learning Interatomic Potentials as Emerging Tools for Materials Science. \emph{Advanced Materials} \textbf{2019}, \emph{31}, 1902765\relax
\mciteBstWouldAddEndPuncttrue
\mciteSetBstMidEndSepPunct{\mcitedefaultmidpunct}
{\mcitedefaultendpunct}{\mcitedefaultseppunct}\relax
\EndOfBibitem
\bibitem[Sch\"{u}tt \latin{et~al.}(2021)Sch\"{u}tt, Unke, and Gastegger]{painn}
Sch\"{u}tt,~K.~T.; Unke,~O.~T.; Gastegger,~M. Equivariant message passing for the prediction of tensorial properties and molecular spectra. 2021; \url{https://arxiv.org/abs/2102.03150}\relax
\mciteBstWouldAddEndPuncttrue
\mciteSetBstMidEndSepPunct{\mcitedefaultmidpunct}
{\mcitedefaultendpunct}{\mcitedefaultseppunct}\relax
\EndOfBibitem
\bibitem[Th{\"o}lke and Fabritiis(2022)Th{\"o}lke, and Fabritiis]{et}
Th{\"o}lke,~P.; Fabritiis,~G.~D. Equivariant Transformers for Neural Network based Molecular Potentials. International Conference on Learning Representations. 2022\relax
\mciteBstWouldAddEndPuncttrue
\mciteSetBstMidEndSepPunct{\mcitedefaultmidpunct}
{\mcitedefaultendpunct}{\mcitedefaultseppunct}\relax
\EndOfBibitem
\bibitem[Batzner \latin{et~al.}(2022)Batzner, Musaelian, Sun, Geiger, Mailoa, Kornbluth, Molinari, Smidt, and Kozinsky]{nequip}
Batzner,~S.; Musaelian,~A.; Sun,~L.; Geiger,~M.; Mailoa,~J.~P.; Kornbluth,~M.; Molinari,~N.; Smidt,~T.~E.; Kozinsky,~B. E(3)-equivariant graph neural networks for data-efficient and accurate interatomic potentials. \emph{Nature Communications} \textbf{2022}, \emph{13}\relax
\mciteBstWouldAddEndPuncttrue
\mciteSetBstMidEndSepPunct{\mcitedefaultmidpunct}
{\mcitedefaultendpunct}{\mcitedefaultseppunct}\relax
\EndOfBibitem
\bibitem[Musaelian \latin{et~al.}(2023)Musaelian, Batzner, Johansson, Sun, Owen, Kornbluth, and Kozinsky]{allegro}
Musaelian,~A.; Batzner,~S.; Johansson,~A.; Sun,~L.; Owen,~C.~J.; Kornbluth,~M.; Kozinsky,~B. Learning local equivariant representations for large-scale atomistic dynamics. \emph{Nature Communications} \textbf{2023}, \emph{14}\relax
\mciteBstWouldAddEndPuncttrue
\mciteSetBstMidEndSepPunct{\mcitedefaultmidpunct}
{\mcitedefaultendpunct}{\mcitedefaultseppunct}\relax
\EndOfBibitem
\bibitem[Batatia \latin{et~al.}(2022)Batatia, Kovacs, Simm, Ortner, and Csanyi]{mace}
Batatia,~I.; Kovacs,~D.~P.; Simm,~G. N.~C.; Ortner,~C.; Csanyi,~G. {MACE}: Higher Order Equivariant Message Passing Neural Networks for Fast and Accurate Force Fields. Advances in Neural Information Processing Systems. 2022\relax
\mciteBstWouldAddEndPuncttrue
\mciteSetBstMidEndSepPunct{\mcitedefaultmidpunct}
{\mcitedefaultendpunct}{\mcitedefaultseppunct}\relax
\EndOfBibitem
\bibitem[Duval \latin{et~al.}(2023)Duval, Mathis, Joshi, Schmidt, Miret, Malliaros, Cohen, Liò, Bengio, and Bronstein]{hitchhiker}
Duval,~A.; Mathis,~S.~V.; Joshi,~C.~K.; Schmidt,~V.; Miret,~S.; Malliaros,~F.~D.; Cohen,~T.; Liò,~P.; Bengio,~Y.; Bronstein,~M. A Hitchhiker's Guide to Geometric GNNs for 3D Atomic Systems. 2023; \url{https://arxiv.org/abs/2312.07511}\relax
\mciteBstWouldAddEndPuncttrue
\mciteSetBstMidEndSepPunct{\mcitedefaultmidpunct}
{\mcitedefaultendpunct}{\mcitedefaultseppunct}\relax
\EndOfBibitem
\bibitem[Eastman \latin{et~al.}(2023)Eastman, Behara, Dotson, Galvelis, Herr, Horton, Mao, Chodera, Pritchard, Wang, De~Fabritiis, and Markland]{spice}
Eastman,~P.; Behara,~P.~K.; Dotson,~D.~L.; Galvelis,~R.; Herr,~J.~E.; Horton,~J.~T.; Mao,~Y.; Chodera,~J.~D.; Pritchard,~B.~P.; Wang,~Y.; De~Fabritiis,~G.; Markland,~T.~E. SPICE, A Dataset of Drug-like Molecules and Peptides for Training Machine Learning Potentials. \emph{Scientific Data} \textbf{2023}, \emph{10}\relax
\mciteBstWouldAddEndPuncttrue
\mciteSetBstMidEndSepPunct{\mcitedefaultmidpunct}
{\mcitedefaultendpunct}{\mcitedefaultseppunct}\relax
\EndOfBibitem
\bibitem[Kovács \latin{et~al.}(2023)Kovács, Moore, Browning, Batatia, Horton, Kapil, Witt, Magdău, Cole, and Csányi]{mace-off}
Kovács,~D.~P.; Moore,~J.~H.; Browning,~N.~J.; Batatia,~I.; Horton,~J.~T.; Kapil,~V.; Witt,~W.~C.; Magdău,~I.-B.; Cole,~D.~J.; Csányi,~G. MACE-OFF23: Transferable Machine Learning Force Fields for Organic Molecules. 2023; \url{https://arxiv.org/abs/2312.15211}\relax
\mciteBstWouldAddEndPuncttrue
\mciteSetBstMidEndSepPunct{\mcitedefaultmidpunct}
{\mcitedefaultendpunct}{\mcitedefaultseppunct}\relax
\EndOfBibitem
\bibitem[Stevenson \latin{et~al.}(2019)Stevenson, Jacobson, Zhao, Wu, Maple, Leswing, Harder, and Abel]{schani}
Stevenson,~J.~M.; Jacobson,~L.~D.; Zhao,~Y.; Wu,~C.; Maple,~J.; Leswing,~K.; Harder,~E.; Abel,~R. Schr\"{o}dinger-ANI: An Eight-Element Neural Network Interaction Potential with Greatly Expanded Coverage of Druglike Chemical Space. 2019; \url{https://arxiv.org/abs/1912.05079}\relax
\mciteBstWouldAddEndPuncttrue
\mciteSetBstMidEndSepPunct{\mcitedefaultmidpunct}
{\mcitedefaultendpunct}{\mcitedefaultseppunct}\relax
\EndOfBibitem
\bibitem[Unke \latin{et~al.}(2021)Unke, Chmiela, Gastegger, Sch\"{u}tt, Sauceda, and M\"{u}ller]{spookynet}
Unke,~O.~T.; Chmiela,~S.; Gastegger,~M.; Sch\"{u}tt,~K.~T.; Sauceda,~H.~E.; M\"{u}ller,~K.-R. SpookyNet: Learning force fields with electronic degrees of freedom and nonlocal effects. \emph{Nature Communications} \textbf{2021}, \emph{12}\relax
\mciteBstWouldAddEndPuncttrue
\mciteSetBstMidEndSepPunct{\mcitedefaultmidpunct}
{\mcitedefaultendpunct}{\mcitedefaultseppunct}\relax
\EndOfBibitem
\bibitem[Anstine \latin{et~al.}(2023)Anstine, Zubatyuk, and Isayev]{aimnet2}
Anstine,~D.; Zubatyuk,~R.; Isayev,~O. AIMNet2: A Neural Network Potential to Meet your Neutral, Charged, Organic, and Elemental-Organic Needs. \textbf{2023}, \relax
\mciteBstWouldAddEndPunctfalse
\mciteSetBstMidEndSepPunct{\mcitedefaultmidpunct}
{}{\mcitedefaultseppunct}\relax
\EndOfBibitem
\bibitem[Gong \latin{et~al.}(2024)Gong, Zhang, Mu, Pu, Wang, Yu, Chen, Zheng, Wang, Chen, Wu, Shi, Gao, Yan, and Xiang]{bamboo}
Gong,~S.; Zhang,~Y.; Mu,~Z.; Pu,~Z.; Wang,~H.; Yu,~Z.; Chen,~M.; Zheng,~T.; Wang,~Z.; Chen,~L.; Wu,~X.; Shi,~S.; Gao,~W.; Yan,~W.; Xiang,~L. BAMBOO: a predictive and transferable machine learning force field framework for liquid electrolyte development. 2024; \url{https://arxiv.org/abs/2404.07181}\relax
\mciteBstWouldAddEndPuncttrue
\mciteSetBstMidEndSepPunct{\mcitedefaultmidpunct}
{\mcitedefaultendpunct}{\mcitedefaultseppunct}\relax
\EndOfBibitem
\bibitem[Ghasemi \latin{et~al.}(2015)Ghasemi, Hofstetter, Saha, and Goedecker]{cent1}
Ghasemi,~S.~A.; Hofstetter,~A.; Saha,~S.; Goedecker,~S. Interatomic potentials for ionic systems with density functional accuracy based on charge densities obtained by a neural network. \emph{Phys. Rev. B} \textbf{2015}, \emph{92}, 045131\relax
\mciteBstWouldAddEndPuncttrue
\mciteSetBstMidEndSepPunct{\mcitedefaultmidpunct}
{\mcitedefaultendpunct}{\mcitedefaultseppunct}\relax
\EndOfBibitem
\bibitem[Faraji \latin{et~al.}(2017)Faraji, Ghasemi, Rostami, Rasoulkhani, Schaefer, Goedecker, and Amsler]{cent2}
Faraji,~S.; Ghasemi,~S.~A.; Rostami,~S.; Rasoulkhani,~R.; Schaefer,~B.; Goedecker,~S.; Amsler,~M. High accuracy and transferability of a neural network potential through charge equilibration for calcium fluoride. \emph{Phys. Rev. B} \textbf{2017}, \emph{95}, 104105\relax
\mciteBstWouldAddEndPuncttrue
\mciteSetBstMidEndSepPunct{\mcitedefaultmidpunct}
{\mcitedefaultendpunct}{\mcitedefaultseppunct}\relax
\EndOfBibitem
\bibitem[Ko \latin{et~al.}(2021)Ko, Finkler, Goedecker, and Behler]{4g}
Ko,~T.~W.; Finkler,~J.~A.; Goedecker,~S.; Behler,~J. A fourth-generation high-dimensional neural network potential with accurate electrostatics including non-local charge transfer. \emph{Nature Communications} \textbf{2021}, \emph{12}\relax
\mciteBstWouldAddEndPuncttrue
\mciteSetBstMidEndSepPunct{\mcitedefaultmidpunct}
{\mcitedefaultendpunct}{\mcitedefaultseppunct}\relax
\EndOfBibitem
\bibitem[Shaidu \latin{et~al.}(2024)Shaidu, Pellegrini, K\"{u}\c{c}\"{u}kbenli, Lot, and de~Gironcoli]{Shaidu2024}
Shaidu,~Y.; Pellegrini,~F.; K\"{u}\c{c}\"{u}kbenli,~E.; Lot,~R.; de~Gironcoli,~S. Incorporating long-range electrostatics in neural network potentials via variational charge equilibration from shortsighted ingredients. \emph{npj Computational Materials} \textbf{2024}, \emph{10}\relax
\mciteBstWouldAddEndPuncttrue
\mciteSetBstMidEndSepPunct{\mcitedefaultmidpunct}
{\mcitedefaultendpunct}{\mcitedefaultseppunct}\relax
\EndOfBibitem
\bibitem[Zubatyuk \latin{et~al.}(2021)Zubatyuk, Smith, Nebgen, Tretiak, and Isayev]{aimnet-nse}
Zubatyuk,~R.; Smith,~J.~S.; Nebgen,~B.~T.; Tretiak,~S.; Isayev,~O. Teaching a neural network to attach and detach electrons from molecules. \emph{Nature Communications} \textbf{2021}, \emph{12}\relax
\mciteBstWouldAddEndPuncttrue
\mciteSetBstMidEndSepPunct{\mcitedefaultmidpunct}
{\mcitedefaultendpunct}{\mcitedefaultseppunct}\relax
\EndOfBibitem
\bibitem[Boittier \latin{et~al.}(2024)Boittier, T\"{o}pfer, Devereux, and Meuwly]{Boittier2024}
Boittier,~E.; T\"{o}pfer,~K.; Devereux,~M.; Meuwly,~M. Kernel-Based Minimal Distributed Charges: A Conformationally Dependent ESP-Model for Molecular Simulations. \emph{Journal of Chemical Theory and Computation} \textbf{2024}, \relax
\mciteBstWouldAddEndPunctfalse
\mciteSetBstMidEndSepPunct{\mcitedefaultmidpunct}
{}{\mcitedefaultseppunct}\relax
\EndOfBibitem
\bibitem[Rappe and Goddard(1991)Rappe, and Goddard]{chargeeq}
Rappe,~A.~K.; Goddard,~W.~A. Charge equilibration for molecular dynamics simulations. \emph{The Journal of Physical Chemistry} \textbf{1991}, \emph{95}, 3358–3363\relax
\mciteBstWouldAddEndPuncttrue
\mciteSetBstMidEndSepPunct{\mcitedefaultmidpunct}
{\mcitedefaultendpunct}{\mcitedefaultseppunct}\relax
\EndOfBibitem
\bibitem[Gao and Remsing(2022)Gao, and Remsing]{scfnn}
Gao,~A.; Remsing,~R.~C. Self-consistent determination of long-range electrostatics in neural network potentials. \emph{Nature Communications} \textbf{2022}, \emph{13}\relax
\mciteBstWouldAddEndPuncttrue
\mciteSetBstMidEndSepPunct{\mcitedefaultmidpunct}
{\mcitedefaultendpunct}{\mcitedefaultseppunct}\relax
\EndOfBibitem
\bibitem[Cheng(2024)]{les}
Cheng,~B. Latent Ewald summation for machine learning of long-range interactions. 2024; \url{https://arxiv.org/abs/2408.15165}\relax
\mciteBstWouldAddEndPuncttrue
\mciteSetBstMidEndSepPunct{\mcitedefaultmidpunct}
{\mcitedefaultendpunct}{\mcitedefaultseppunct}\relax
\EndOfBibitem
\bibitem[Kim \latin{et~al.}(2024)Kim, King, Zhong, and Cheng]{cacelr}
Kim,~D.; King,~D.~S.; Zhong,~P.; Cheng,~B. Learning charges and long-range interactions from energies and forces. 2024; \url{https://arxiv.org/abs/2412.15455}\relax
\mciteBstWouldAddEndPuncttrue
\mciteSetBstMidEndSepPunct{\mcitedefaultmidpunct}
{\mcitedefaultendpunct}{\mcitedefaultseppunct}\relax
\EndOfBibitem
\bibitem[Simeon and De~Fabritiis(2023)Simeon, and De~Fabritiis]{tensornet}
Simeon,~G.; De~Fabritiis,~G. TensorNet: Cartesian Tensor Representations for Efficient Learning of Molecular Potentials. Advances in Neural Information Processing Systems. 2023; pp 37334--37353\relax
\mciteBstWouldAddEndPuncttrue
\mciteSetBstMidEndSepPunct{\mcitedefaultmidpunct}
{\mcitedefaultendpunct}{\mcitedefaultseppunct}\relax
\EndOfBibitem
\bibitem[Pelaez \latin{et~al.}(2024)Pelaez, Simeon, Galvelis, Mirarchi, Eastman, Doerr, Th\"{o}lke, Markland, and De~Fabritiis]{tmdnet}
Pelaez,~R.~P.; Simeon,~G.; Galvelis,~R.; Mirarchi,~A.; Eastman,~P.; Doerr,~S.; Th\"{o}lke,~P.; Markland,~T.~E.; De~Fabritiis,~G. TorchMD-Net 2.0: Fast Neural Network Potentials for Molecular Simulations. 2024; \url{https://arxiv.org/abs/2402.17660}\relax
\mciteBstWouldAddEndPuncttrue
\mciteSetBstMidEndSepPunct{\mcitedefaultmidpunct}
{\mcitedefaultendpunct}{\mcitedefaultseppunct}\relax
\EndOfBibitem
\bibitem[Bannwarth \latin{et~al.}(2019)Bannwarth, Ehlert, and Grimme]{xtb}
Bannwarth,~C.; Ehlert,~S.; Grimme,~S. GFN2-xTB—An Accurate and Broadly Parametrized Self-Consistent Tight-Binding Quantum Chemical Method with Multipole Electrostatics and Density-Dependent Dispersion Contributions. \emph{Journal of Chemical Theory and Computation} \textbf{2019}, \emph{15}, 1652–1671\relax
\mciteBstWouldAddEndPuncttrue
\mciteSetBstMidEndSepPunct{\mcitedefaultmidpunct}
{\mcitedefaultendpunct}{\mcitedefaultseppunct}\relax
\EndOfBibitem
\bibitem[Artrith \latin{et~al.}(2011)Artrith, Morawietz, and Behler]{3g1}
Artrith,~N.; Morawietz,~T.; Behler,~J. High-dimensional neural-network potentials for multicomponent systems: Applications to zinc oxide. \emph{Phys. Rev. B} \textbf{2011}, \emph{83}, 153101\relax
\mciteBstWouldAddEndPuncttrue
\mciteSetBstMidEndSepPunct{\mcitedefaultmidpunct}
{\mcitedefaultendpunct}{\mcitedefaultseppunct}\relax
\EndOfBibitem
\bibitem[Morawietz \latin{et~al.}(2012)Morawietz, Sharma, and Behler]{3g2}
Morawietz,~T.; Sharma,~V.; Behler,~J. A neural network potential-energy surface for the water dimer based on environment-dependent atomic energies and charges. \emph{The Journal of Chemical Physics} \textbf{2012}, \emph{136}\relax
\mciteBstWouldAddEndPuncttrue
\mciteSetBstMidEndSepPunct{\mcitedefaultmidpunct}
{\mcitedefaultendpunct}{\mcitedefaultseppunct}\relax
\EndOfBibitem
\bibitem[Larsen \latin{et~al.}(2017)Larsen, Mortensen, Blomqvist, Castelli, Christensen, Dułak, Friis, Groves, Hammer, Hargus, Hermes, Jennings, Jensen, Kermode, Kitchin, Kolsbjerg, Kubal, Kaasbjerg, Lysgaard, Maronsson, Maxson, Olsen, Pastewka, Peterson, Rostgaard, Schiøtz, Schütt, Strange, Thygesen, Vegge, Vilhelmsen, Walter, Zeng, and Jacobsen]{ase-paper}
Larsen,~A.~H.; Mortensen,~J.~J.; Blomqvist,~J.; Castelli,~I.~E.; Christensen,~R.; Dułak,~M.; Friis,~J.; Groves,~M.~N.; Hammer,~B.; Hargus,~C.; Hermes,~E.~D.; Jennings,~P.~C.; Jensen,~P.~B.; Kermode,~J.; Kitchin,~J.~R.; Kolsbjerg,~E.~L.; Kubal,~J.; Kaasbjerg,~K.; Lysgaard,~S.; Maronsson,~J.~B.; Maxson,~T.; Olsen,~T.; Pastewka,~L.; Peterson,~A.; Rostgaard,~C.; Schiøtz,~J.; Schütt,~O.; Strange,~M.; Thygesen,~K.~S.; Vegge,~T.; Vilhelmsen,~L.; Walter,~M.; Zeng,~Z.; Jacobsen,~K.~W. The atomic simulation environment—a Python library for working with atoms. \emph{Journal of Physics: Condensed Matter} \textbf{2017}, \emph{29}, 273002\relax
\mciteBstWouldAddEndPuncttrue
\mciteSetBstMidEndSepPunct{\mcitedefaultmidpunct}
{\mcitedefaultendpunct}{\mcitedefaultseppunct}\relax
\EndOfBibitem
\bibitem[Bahn and Jacobsen(2002)Bahn, and Jacobsen]{ase-ref-2}
Bahn,~S.~R.; Jacobsen,~K.~W. An object-oriented scripting interface to a legacy electronic structure code. \emph{Comput. Sci. Eng.} \textbf{2002}, \emph{4}, 56--66\relax
\mciteBstWouldAddEndPuncttrue
\mciteSetBstMidEndSepPunct{\mcitedefaultmidpunct}
{\mcitedefaultendpunct}{\mcitedefaultseppunct}\relax
\EndOfBibitem
\bibitem[Schwilk \latin{et~al.}(2020)Schwilk, Tahchieva, and von Lilienfeld]{qmspin}
Schwilk,~M.; Tahchieva,~D.~N.; von Lilienfeld,~O.~A. Large yet bounded: Spin gap ranges in carbenes. 2020; \url{https://arxiv.org/abs/2004.10600}\relax
\mciteBstWouldAddEndPuncttrue
\mciteSetBstMidEndSepPunct{\mcitedefaultmidpunct}
{\mcitedefaultendpunct}{\mcitedefaultseppunct}\relax
\EndOfBibitem
\end{mcitethebibliography}
